# Matching Edges in Images ; Application to Face Recognition

Joël Le Roux*, Philippe Chaurand** and Mickaël Urrutia**,

*Professor, **Students,
Polytechnic Engineering School of the University of Nice,
Department of Computer Science and Engineering
930  Route des Colles, BP 145
06903 Sophia Antipolis, France

Contact : leroux@polytech.unice.fr

22 March 2006

*Abstract* : This communication describes a representation of images as a set of edges characterized by their position and orientation. This representation allows the comparison of two images and the computation of their similarity. The first step in this computation of similarity is the seach of a geometrical basis of the two dimensional space where the two images are represented simultaneously after transformation of one of them.
Presently, this simultaneous representation takes into account a shift and a scaling ; it may be extended to rotations or other global geometrical transformations. An elementary probabilistic computation shows that a sufficient but not excessive number of trials (a few tens) ensures that the exhibition of this common basis is guaranteed in spite of possible errors in the detection of edges. When this first step is performed, the search of similarity between the two images reduces to counting the coincidence of edges in the two images. The approach may be applied  to many problems of pattern matching ; it was checked  on face recognition.

### 1. Introduction

In many types of images, edges and contours are probably the most important informative elements. Many processing methods are based on the detection of edges and their representation as chains of edges or contours. For a good reference on this kind of approach, in a context similar to that of our work, we suggest the paper of  Y. Gao and M. K. H. Leung [1]. General references on face recognition may be found in [2]. This chaining is often a source of difficulties due to detection failures. However this step is not really necessary in several applications where it is sufficient to know the position and the orientation of unchained edges ; this information is very redundant and is often used for chaining,  but it can also be used directly, and thus avoiding a number of drawbacks. The fact that we base the search on very elementary data as edges instead of grouping them in lines differenciates our work from that of Gao and Leung. These edges are the data that we use in the search of a match between two images, taking into account the fact that some edges may be missing, or in excess in one of the two images ; it is also possible that they are poorly identified in terms of position and orientation when the contour is not sharp in the image. Once these edges are detected and characterized, as we look for a correspondence between the two images, it is interesting to find a two dimensional space basis where both images can be represented. In most cases, two (sometimes three) edges  are sufficient to define a basis in an image ; and when two couples of edges are matched in the two images, they can be used to contruct a common basis of the images, or the geometrical transformation of one image into the other.

In general, one cannot ensure *a priori* that the two couples of edges are correctly paired, but most of the time, this can be checked *a posteriori* : indeed, when the pairing is correct and when there is a correspondence between the images, it is possible to verify that a large proportion of edges coincide in the two images, the difference being only due to the elements that are missing or in excess ; when the pairing is not correct, or when there is no similarity between the two images, the proportion of coincidence will be sensibly reduced. These remarks will be used in the search of a correspondence between the two images.

*Content of the communication* : The second section describes the different steps of the search of a correspondence : detection of edges, search of a possible common basis and *a posteriori* verification of the validity of this common basis, completed by a justification. The third section gives an illustration of the approach applied to the matching of faces images. Finally a short discussion on the interest and drawbacks of the method is given in the conclusion.

## 2. The Steps of the Matching Process

The first paragraph describes the search of edges ; the second describes the parameters used in the characterization of a basis ; the third describes the method used in order to exhibit couples of edges that can serve in the construction of a common basis of the two images ; the fourth describes the verification methods ensuring that the basis choice is pertinent and where the matching score is updated ; the fifth proposes a probabilistic justification.

### 2.1. The search of edges in an image

A edge is a pixel (or a tiny region of the image) where the gradient is large and where, in many cases, the neighbouring pixels also present a large gradient and an orientation close to that of the first (low curvature). These characteristics can be computed in many different ways that often reduce to more or less elaborated linear filterings, followed by a decision step where the gradient and curvature are compared to thresholds. We have chosen to perform these computations in using a frequency domain representation : instead of approximating a partial derivative with respect to the coordinates $x$ or $y$, one computes the Fourier transform $F(u,v)$ of the image $f(x,y)$ and multiplies it by $j.u$ in order to obtain $j.u\, F(u,v)$, tranform of $\partial f/\partial x$, by $j.v$ to obtain $j.v\, F(u,v)$ transform of $\partial f/\partial y$ ; thus, one gets the *exact value* of the gradient of the image before sampling (continuous function of the spatial variables $x$ and $y$, when one assumes that the Shannon conditions are satisfied.) In doing so, one can easily include a bidimensional linear filter in order to reduce the effect of noise ; one can also compute the second derivatives and the curvature, avoiding difficulties inherent to the sampling process ; however, the proposed approach is not specially sensitive to the method used for the gradient computation provided that the images are not very noisy and that the computation method yields a correct estimation of the direction of a edge (orthogonal to the gradient direction) ; this direction is an angle on a $2\pi$ interval and not a $\pi$ interval.

In a first selection, we only consider the edges with large gradient and low curvature: it is among these elements whose detection is considered as reliable, that we search those likely to be efficiently used in a basis construction ; but other elements, considered as less reliable can be used in the following procedure where the validity of the basis is checked. In doing so, we have, for an image, a list of edges giving for each of them its position, its orientation, possibly its curvature and an index of confidence taking into account the gradient amplidude and the noise level.

## 2.2. Definition of a basis from edges

In order to define a basis in an image, we need two noncolinear vectors. In general two edges that are sufficiently distincts can be used for such a definition. In some cases (edges on straight lines), three edges are needed, but this does not modify the method. It is possible to reduce the number of possibilities in the search of a correspondence between the two images, in accepting some constraints on the couple of elements likely to be elected as basis. For example, one can assume that the first edge is horizontal in the upper part of the image and that the second is vertical on the left. This kind of restriction has no consequence on the method, but it may have consequences on the result : in adding constraints, we can miss the good match between two images ; reducing them may yield to an important increase of the number of coincidences to be checked in order to ensure the validity of the basis. In our experiments we have chosen to select edges with low curvature, rather distant from one another and sufficiently sharp (low detection error risk due to the noise on the image).

## 2.3. Couples of edges eligible for the construction of a common basis of the two images

In order to simplify the presentation, and in avoiding the treatment of more complex cases, we only consider a shift and a scaling of the image ; the extension to rotations and simple transformations as anisotropic scalings is direct. We have two images : the old one named $A$ that is generally memorized in a database of reference images and has a good quality ; and the new one $N$ for which we look for a possible correspondent in the database while edges may be missing or in excess.

### 2.3.1. Parameters characterizing a basis

In the image $A$, we select two edges ($EG$) distinct enough and sufficiently well detected (we are sure that both are significant edges) ; for example, one is close to horizontal in the upper part of the image and the other close to vertical in the middle left part of the image ; these two edges, that we name $EG_{A1}$ and $EG_{A2}$ will serve as a first hypothesis in the search of a common basis for $A$ and $N$. The characteristics of these two edges are their coordinates $x_{A1}$ and $x_{A2}$, $y_{A1}$ and $y_{A2}$ and their orientations $\theta_{A1}$ and $\theta_{A2}$. From the coordinates of the edges, we can deduce the slope $\varphi_{A12}$ of the axis connecting them (fig. 1).

### 2.3.2. Search in the new image of a couple of edges susceptible of coinciding with the basis chosen in the old one

Now, we have to find in image $N$, two edges $EG_{N1}$ and $EG_{N2}$, that correspond to the couple ($EG_{A1}$, $EG_{A2}$) after a possible translation and scaling. We select in $N$ an edge with slope close to $\theta_{A1}$, that we name $EG_{N1}$, and we look if there exists an edge with slope $\theta_{A2}$, in the direction $\varphi_{A12}$, starting from the point of coordinates ($x_{N1}$, $y_{N1}$). If so, this gives us the second edge, that we name $EG_{N2}$. The distance between $EG_{N1}$ and $EG_{N2}$ allows the computation of the scaling factor for the transformation from the basis of $N$ into the basis of $A$. The distance between angles takes into account the uncertainty on the measures of the characteristics of the edges. Of course, this search may fail for two kinds of reasons : either there is no correspondence between the two images, or there are errors in the edges detection procedure. We shall discuss this point in paragraph 2.5. On the other hand, several compatible couples can be found in $N$. Their acceptation or rejection will occur in the subsequent verification procedure.

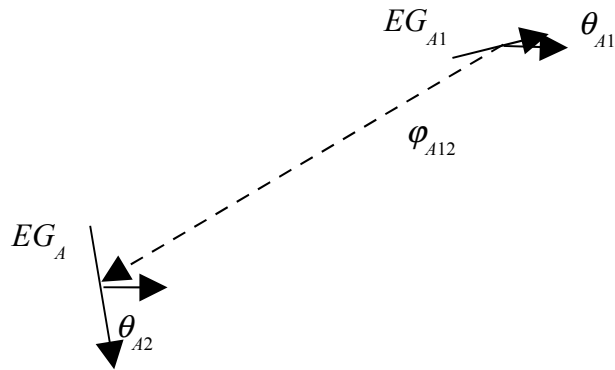

Figure 1. Parameters of the two edges used to define a basis

For the present, in case of failure of this search, we start over again the search of a couple ($EG_{N1}$, $EG_{N2}$) several times (some tens) ; if all these searches fail, we start over again a new search based on another couple ($EG_{A1}$, $EG_{A2}$) in image $A$. If, after a certain number (some tens) of trials of this type, the search has failed, it is reasonable to conclude that it is not possible to find a correspondence between the two images. Now we assume that we have found a couple ($EG_{N1}$, $EG_{N2}$) corresponding to a couple ($EG_{A1}$, $EG_{A2}$) (fig. 2), and we have to confirm or invalidate this hypothesis.

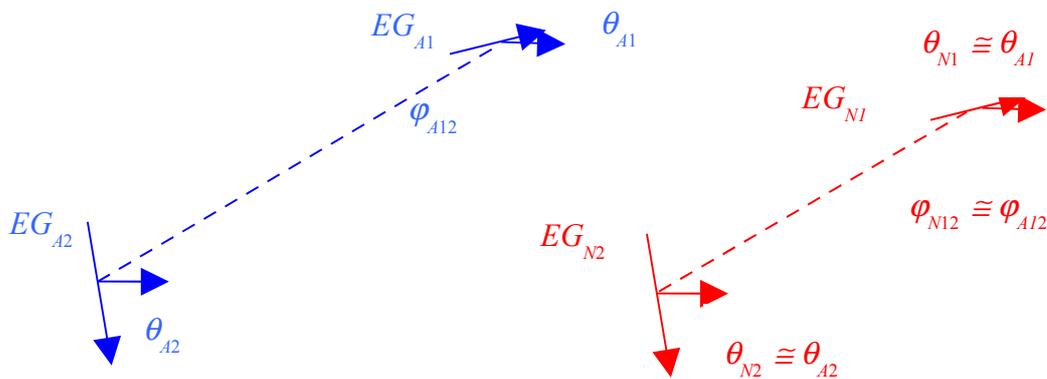

Figure 2. Correspondence of two edges in the two images ;
''equalities between angles'' or coordinates are approximative :
they take into account the possible errors on the measurements of the
characteristics of edges.

If both edges are on straight lines, the correspondence is not unique, and a third edge is necessary in order to define a common basis for the two images ; the search of the correspondence on the third edge is included in the following step : verification of the validity of the basis.

## 2.4. Vérification *a posteriori* of the validity of the correspondence

If we have found in the two images two compatible couples $(EG_{A1}, EG_{A2})$, $(EG_{N1}, EG_{N2})$, the correspondence between the two images is possible, but this has to be verified. Our basical assumption is the following : if there is actually a correspondence, we will find in the two images a significant number of edges that have a correspondent in the other (same position and same orientation, after basis change in one image). It is possible that a match of the two basis is accepted, although it is not the correct one ; in that case the number of coincidences between edges in the two images will remain low, just as in the case where there is no correspondence between the two images. After applying to $N$ the transformation that changes the basis $(EG_{N1}, EG_{N2})$ into $(EG_{A1}, EG_{A2})$, we count the number of edges that coincide (same position, same orientation, taking into account the measurement errors) (fig. 3). In the search of couples of edges allowing the construction of a common basis, it is advisable to consider only edges whose detection is reliable, but in the verification procedure, it is possible to take into account edges with a lower confidence index.

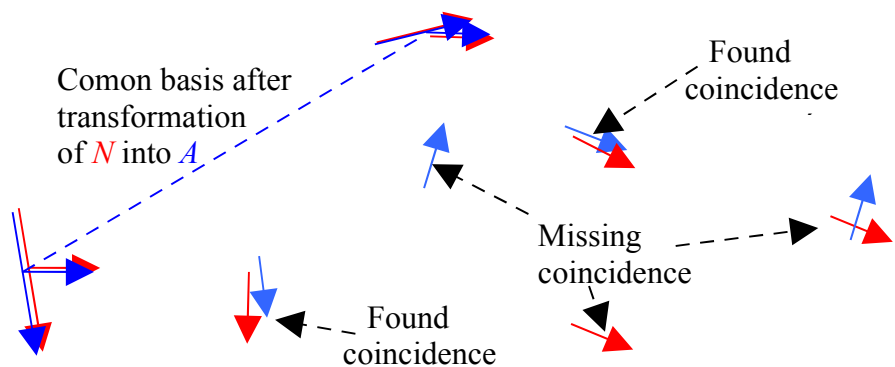

Figure 3. Verification of the pertinence of the choice of a common basis for the two images : we count the number of edges that coincide in the two images ; if this number is large, we say that the two images are similar ; on the contrary, if this number is small, the two images are not similar or the correspondence between basis is erroneous.

## 2.5. Some elements for a justification

The (common sense) basical hypothesis for the justification of the approach is that in the case where there is no correspondence between the images, or where the choice of a common basis is not correct, the proportion of edges that coincide in the two images is small ; on the contrary, if the common basis is correctly chosen, and if there is actually a correspondence between the two images, a significant proportion of edges of the two images coincides ; the lack of coincidences being due to errors in the edges detection process. In the justification of our approach, the main problem is to show that, most of the times, it will be possible to find a common basis for the two images *A* and *N* when they coincide effectively in spite of possible errors in the detection of edges.

If we assume that the quality of image *A* and its representation as a set of edges are good, the source of errors is the search of a basis is the absence of detection of a edge in *N*. The error related to the presence of an edge in excess in *N* does not raise this kind of difficulty : it will yield a rejection of the hypothesis in the verification procedure.

We propose a probabilistic justification (under simplifying hypotheses) in order to evaluate the risk that a common basis is missed although the two images actually coincide. We suppose that an edge considered as significant in *A* is not detected in *N* with probability *p*, although there is a correspondence in the two images. We suppose that in *A*, there are $m_{A1}$ candidates likely to be considered as the first element of a basis couple and $m_{A2}$ candidates likely to be the second ; both number being of the same order of magnitude, we take: $m_{A1} = m_{A2} = m$. The probability that in the edges of *N* that we consider, there is no couple corresponding to two edges of these two groups is about

$$[1-(1-p)^2]^m$$

When *p* is equal to 0.25, the probability that we miss the basis is about $10^{-7}$ when *m* is equal to 20, and about $10^{-9}$ when is equal to 25. If we check a few tens of couples of edges, we are almost certain that we will not reject *N* when there is actually a correspondence between the two images. An estimation of the average number of trials ensuring a correct selection of the couples of edges is given by $(1-p)^{-2}$, that is 16/9 for $p = 0.25$ and about 1.2 for $p = 0.1$. The analysis of the case where three edges are needed (edges on straight lines) yields a comparable although slighly larger result. (about 50 trials are required).

A probabilistic analysis of the verification step does not give pertinent information : it appears that the proportion of edges likely to coincide between random images is very small ; in informative images, the proportion of edges that coincide depend on the similarity between the two images and on the penalization of the difference between paired edges or edges close to one another in the two images.

### 3. Application to face matching

The proposed approach may be applied to most problems of pattern matching where the transformation of one image into the other is a simple geometrical transform (shift, scaling, rotation). We have checked its validity in the case of face matching where a face is represented by several hundreds of edges.

**3.1. Implementation**

The parameters of the edges of *A* (position, orientation, quality) are memorized. A few hundreds of couples of edges likely to be a basis are selected. They are ordered according to a criterion taking into account the quality of their detection, their distance and the difference of their orientations. When *N* is analyzed, one looks if correspondents to these couples of *A* are found ; in each case a confidence factor is computed according to the pertinence of the match and the basis transformation is characterized (shift and scaling factor). This corresponds to the separation of the search branches (fig. 4). Then the verification takes place: the presence of a correspondent of a third, a fourth, ... edge of *A* is checked in *N* after its tranformation and the confidence is updated (misses are accepted, but they reduce the confidence factor of the search branch) ; in this verification, lower quality edges may be considered, while only higher quality edges are accepted in the common basis search ; if the confidence becomes lower than a threshold, the branch is abandonned (fig. 4) ; if all branches yield a poor result, then the match fails and we conclude the absence of coincidence between the two images.

### 3.2. Some results

Fig. 5 gives an example of application to face profiles ; in that case a successful match where the scaling factor is 1.12 ; we note that in this example, the criterion of the common basis selection is rather poor and could probably be improved with a better characterization of the selected couple of edges (noise level in the neighbourhood, sharpness, curvature, etc ...) ; fig. 6 gives an example where the match fails.

### 3.3. Some comments ; improvements under consideration

The search of a common basis is efficient if the edges selection technique clearly enhances the most "significant" edges (sharp and not noisy edges, low curvature) ; however, the second step, where the validity of the basis is checked, may use edges of lesser quality : less visible edges that are present in both images with the right position and orientation contribute to the confirmation of the hypothesis of coincidence of the two basis and of the two images. In the verification step, it is possible to focus the analysis on regions of the image where the correspondence is good that is where the the proportion of matching edges is high. It is also possible to use different strategies, for example firstly a coarse seach in order to select possible candidates in an image database followed by a refined search in order to confirm or invalidate an hypothesis.

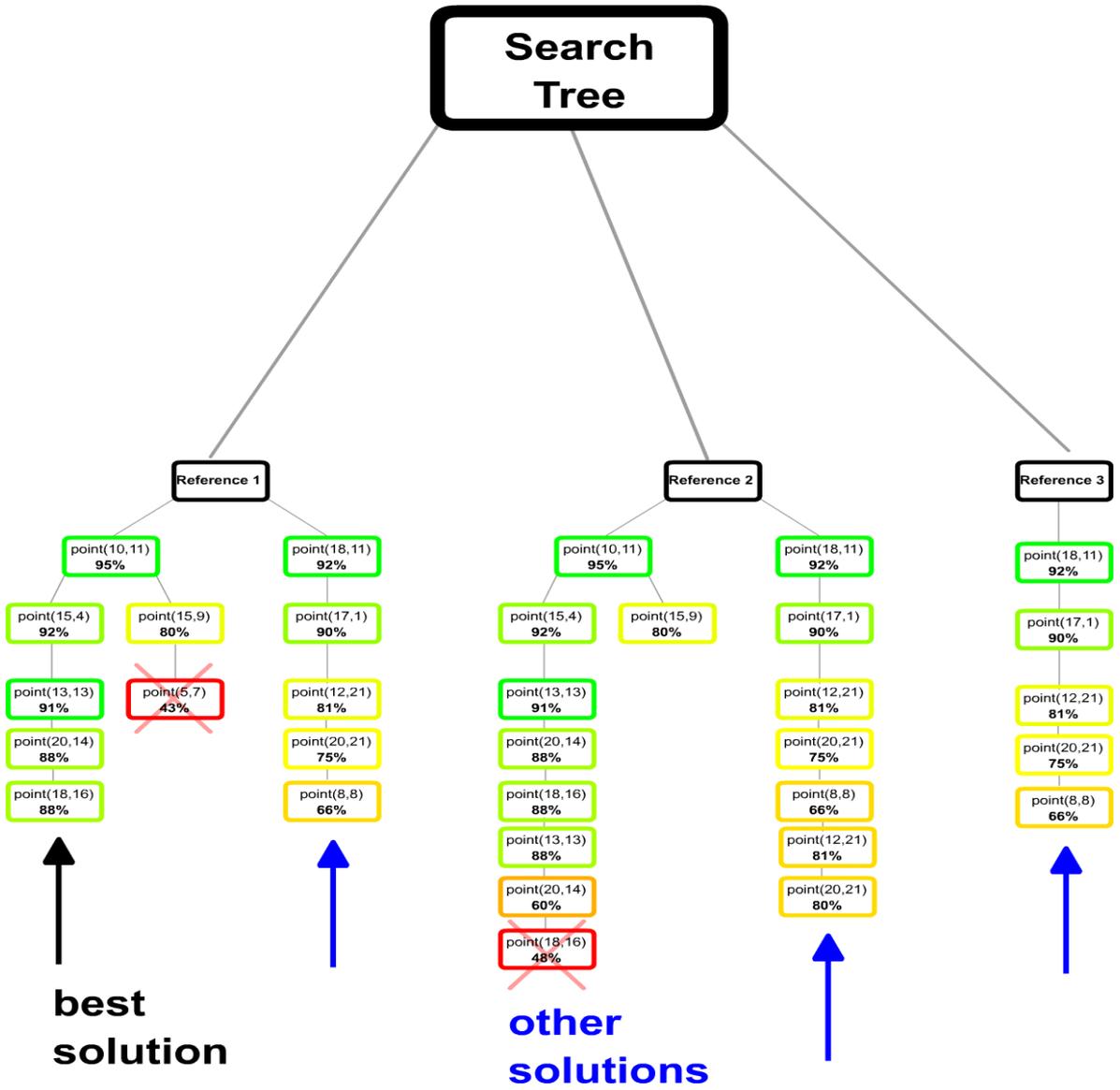

Fig. 4 : Construction of the search tree in the verification process : each branch "reference" correspond to a possible common basis for *A* and *N* and to the subsequent sequence of verification ; when the confidence is too low, the analysis along the branch is cancelled

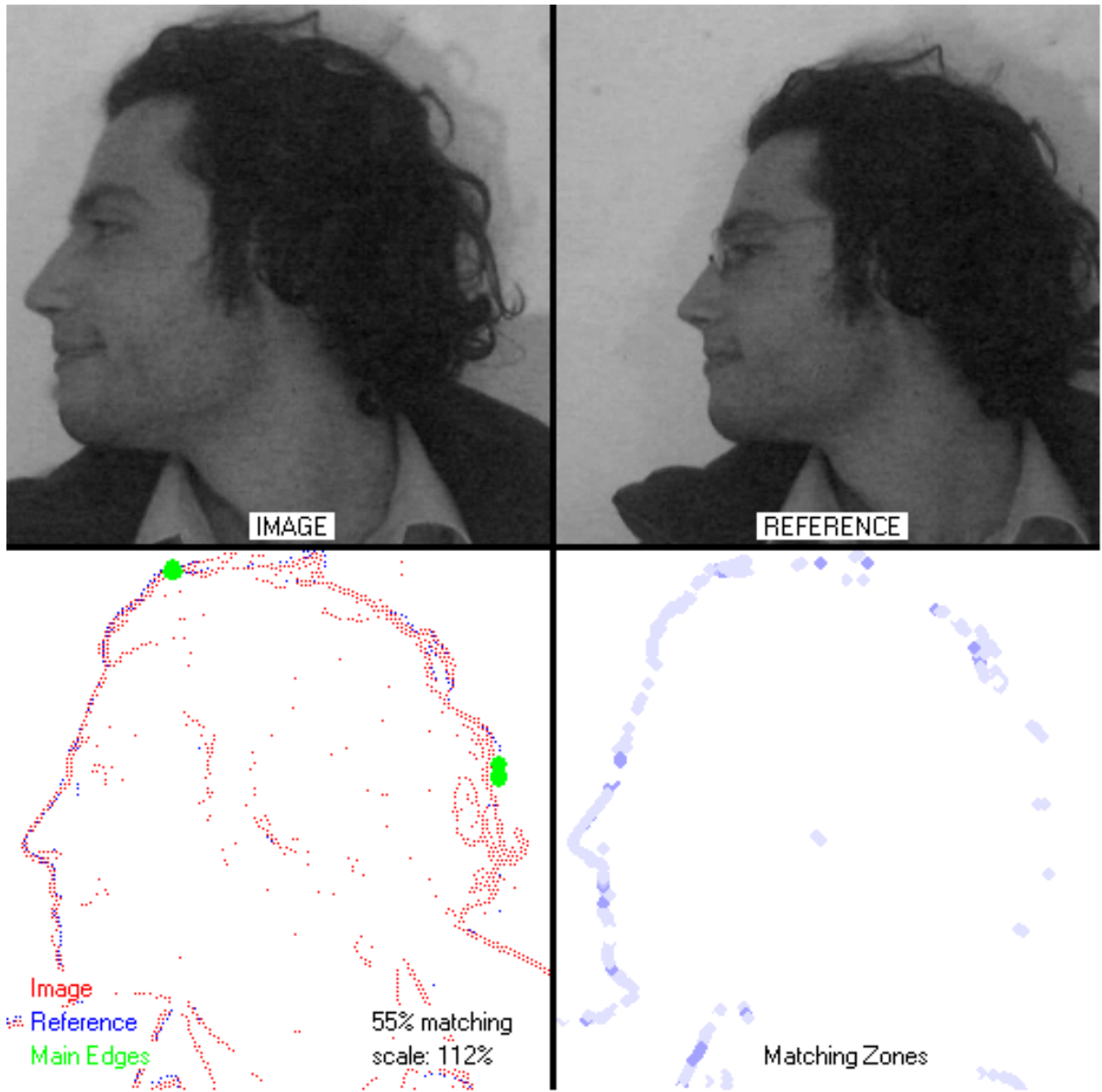

Fig. 5 : Example of match between a model and a new image ; the orientation of the edges is taken into account in matching the two images although not presented on this figure ; in spite of a poor choice of the common basis ("main edges"), there is a reasonable match between the two images.

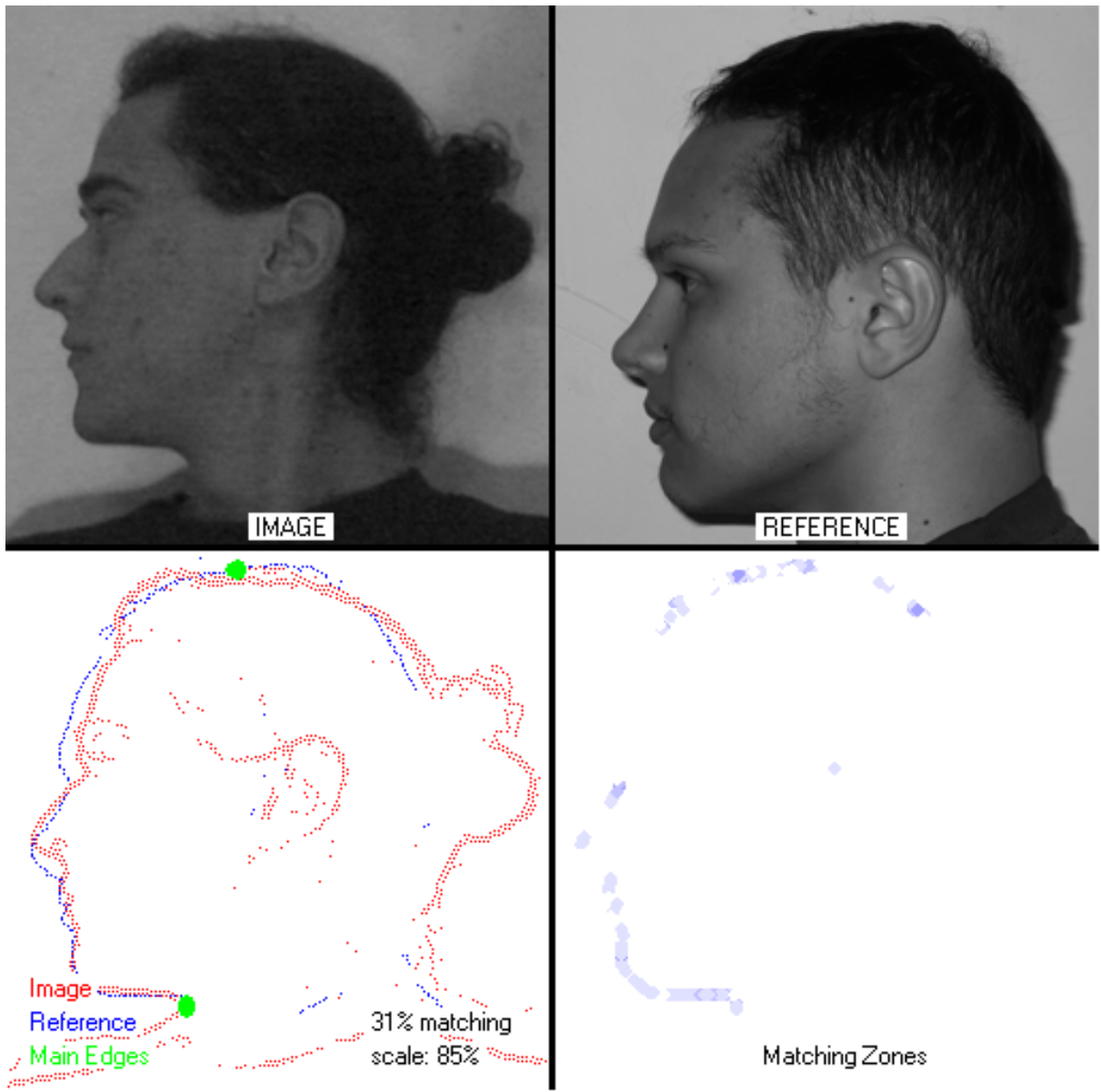

Fig. 6 : Example of bad match between a model and a new image.

## 4. Conclusion

Performing the match on very simple informative elements like edges it presents several advantages related to the important redundancy of the representation : for example if one base the match search on data produced by sophisticated computations, the match may fail due to misinterpretations occuring during these computations. Among the potential interests of the approach is the possibility to apply it to partly hidden objects, to the recognition of a simple object in a complex image ; we have considered simple transformations but it is possible to take into account rotations or more complex global geometrical transformations. Another interesting potential application is the fingerprint analysis where the present approach could be a possible complement to usual approaches. These often look for very informative characteristic points, that may be missed during the analysis process ; while the high degree of redundance and the robustness of the present approach based on edges detection guarantees the match with a high degree of confidence.

Of course, the approach presents a drawback : the amount of computations may be very important ; if one considers the recognition of one object among many,  the amont of computations may be reduced in using strategies classical in tree search ; one can also start with a coarse fast search , where the penalties of errors are reduced, followed by refined strategies applied to similar parts of the image.

Our purpose was not to exhibit excellent performances, but rather to illustrate the fact that a quite simple method could yield interesting results. The software was developped as an initiation in image recognition. The implementation was done in a four monthes (November 2005 February 2006) part time project by last year students of the Computer Science and Engineering Department of the School (France). A report and some slides about this experiment are available [3].   The method may perhaps be used as an efficient complement to other matching techniques.